\documentclass[10pt,twocolumn,letterpaper]{article}

\usepackage{iccv}
\usepackage{times}
\usepackage{epsfig}
\usepackage{graphicx}
\usepackage{amsmath}
\usepackage{amssymb}

\usepackage{cite}
\usepackage{makecell}
\usepackage{ctable}
\usepackage{arydshln}
\usepackage{color}
\usepackage{xcolor}
\usepackage{pifont}
\usepackage{subcaption}
\usepackage{subfiles}
\definecolor{myred}{HTML}{B85450}
\definecolor{myblue}{HTML}{0066CC}
\definecolor{mygreen}{HTML}{009900}
\definecolor{myyellow}{HTML}{D79B00}
\definecolor{myorange}{HTML}{FF8000}

\newcommand\mc[1]{\multicolumn{1}{c}{#1}}
\newcommand{\cdashlinelr}[1]{%
  \noalign{\vskip\aboverulesep
           \global\let\@dashdrawstore\adl@draw
           \global\let\adl@draw\adl@drawiv}
  \cdashline{#1}
  \noalign{\global\let\adl@draw\@dashdrawstore
           \vskip\belowrulesep}}

\usepackage[pagebackref=true,breaklinks=true,letterpaper=true,colorlinks,bookmarks=false]{hyperref}

\iccvfinalcopy 

\begin{document}

\title{Cooperative Embeddings for Instance, Attribute and Category Retrieval}

\author{William Thong \qquad Cees G. M. Snoek \qquad Arnold W. M. Smeulders\\
University of Amsterdam \\
{\tt\small \{w.e.thong, cgmsnoek, a.w.m.smeulders\}@uva.nl}}

\maketitle

\begin{abstract}
The goal of this paper is to retrieve an image based on instance, attribute and category similarity notions.
Different from existing works, which usually address only one of these entities in isolation, we introduce a cooperative embedding to integrate them while preserving their specific level of semantic representation.
An algebraic structure defines a superspace filled with instances. Attributes are axis-aligned to form subspaces, while categories influence the arrangement of similar instances. These relationships enable them to cooperate for their mutual benefits for image retrieval.
We derive a proxy-based softmax embedding loss to learn simultaneously all similarity measures in both superspace and subspaces.
We evaluate our model on datasets from two different domains.
Experiments on image retrieval tasks show the benefits of the cooperative embeddings for modeling multiple image similarities, and for discovering style evolution of instances between- and within-categories.
\end{abstract}

\section{Introduction}

\begin{figure}[t]
\begin{center}
\includegraphics[width=\linewidth]{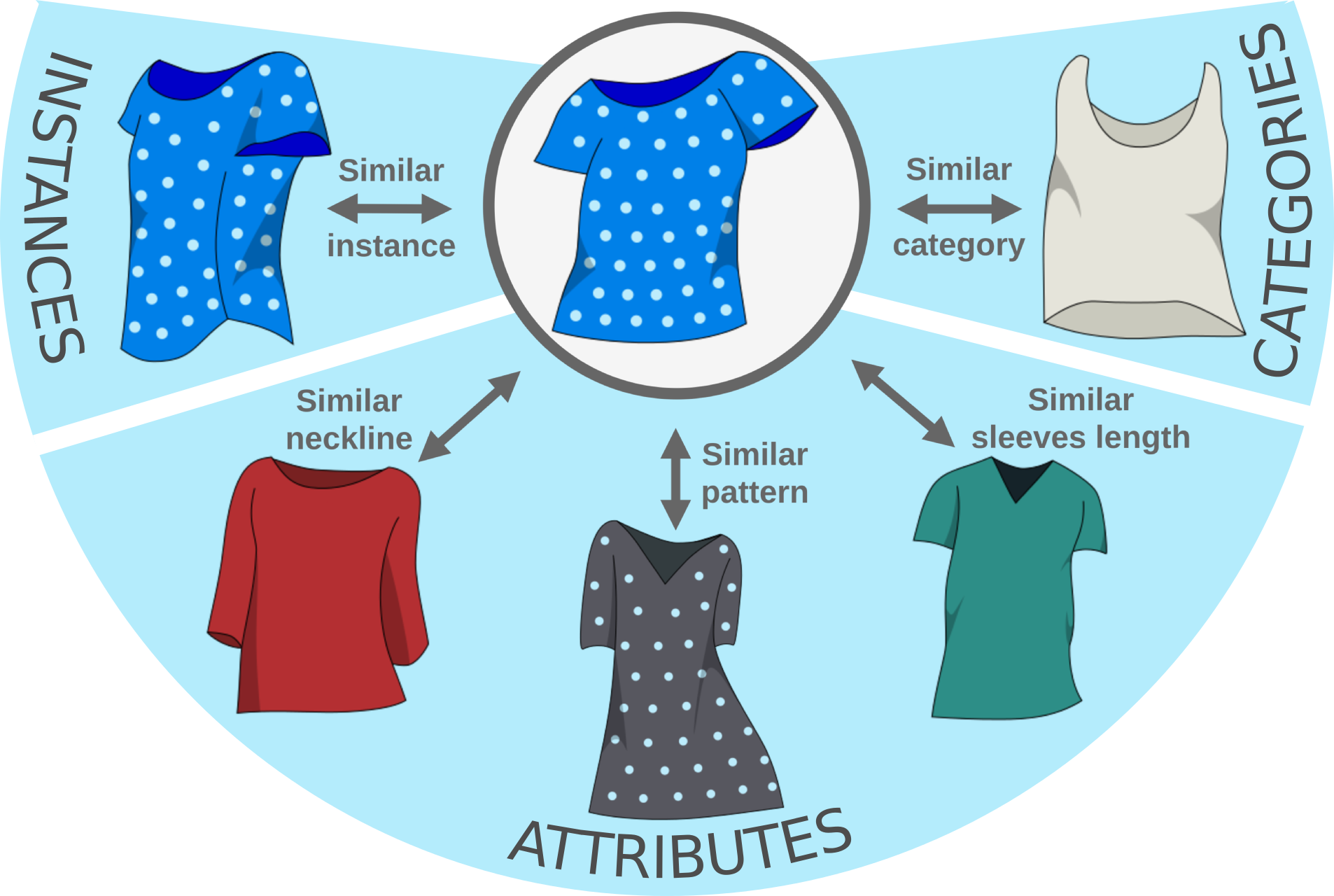}
\end{center}
   \caption{We relate images through multiple similarities for image retrieval, via a single and cooperative embedding. The circled image (center) relates to the others through different levels of semantic representation, \ie instance- (left), attribute- (bottom) and category-related (right) notions.}
\label{fig:one}
\end{figure}

This paper focuses on image retrieval based on instance, attribute and category similarities in \textit{cooperation}. 
There is a considerable body of work on instance retrieval, evaluating similarity between a query and any target instance on the basis of their visual appearance, \eg~\cite{chopra2005learning,schroff2015facenet,songCVPR16,movshovitz2017no};
as well as on attribute retrieval, evaluating the similarity by attribute values, \eg~\cite{finegrained,veit17conditionalsimilarity,zhao2018modulation}; and category retrieval, evaluating common memberships, \eg~\cite{frome2007image,chechik2009online,deselaers2011visual,mensink2013distance}.
From the user's perspective, the desired outcome will, however, be a cooperation between these three different forms of retrieval. Consider Figure~\ref{fig:one}, the attributes of an instance of a \textit{blouse} may share some similarities with the attributes of other \textit{dresses}. In fact, one instance of a category is aligning its similarity to instances of another category through their attribute values. This type of cooperative search expands the individual capacity of instance, attribute or category search by combining their different levels of semantic representation. We propose to integrate these three entities into a single embedding to enable cooperation in search.

A natural approach to combine multiple entities is to rely on multitask learning~\cite{caruana1997multitask}. Combining entities has proven to be effective in various computer vision tasks~\cite{Chen2018,taskonomy2018,kokkinos2017ubernet}. Instead of considering them in isolation, multitask learning builds a shared representation to leverage the benefits coming from combining multiple entities~\cite{ruder2017overview}.
In image retrieval, recent works have explored how to model multiple attribute notions jointly~\cite{veit17conditionalsimilarity,zhao2018modulation}.
Veit~\etal~\cite{veit17conditionalsimilarity} concatenate attribute subspaces while Zhao~\etal~\cite{zhao2018modulation} condition the embedding on attributes via a modulation module. Both show the joint attribute representation to be superior to a set of individual models, one for each attribute. In this paper, we go beyond combining attributes and let them cooperate with instances and categories.

One approach to achieve three-entity combination would be to concatenate instance and category subspaces to the attribute subspaces in~\cite{veit17conditionalsimilarity}. This expands the embedding space without exploiting the underlying semantic structure that relates them. For a series of shirts (instances) getting longer and longer (evolution of attribute values) until it becomes a dress (category change), it would be desirable for a meaningful search to have these three cooperate. Our cooperative embedding enables such a cooperation.
Another approach would be to add instance and category conditions to the modulation in~\cite{zhao2018modulation}. While being very efficient during training, this creates computational complexities by extending the inference time and the storage space needed to extract embeddings of all entities. 
Starting from the idea of attribute subspaces proposed by Veit~\etal~\cite{veit17conditionalsimilarity}, we integrate instance and category similarity notions. We explicitly enforce attributes to compose the visual properties of instances, and similar instances to form categories.

Learning a similarity metric between a search query and relevant images usually relies on Siamese~\cite{chopra2005learning,hadsell2006dimensionality} or triplet strategies~\cite{weinberger2009distance,schroff2015facenet}.
They have led to impressive image retrieval results~\cite{bell2015learning,hermans2017defense,amos2016openface,arandjelovic2016netvlad,songCVPR16,liuLQWTcvpr16DeepFashion,sangkloy2016sketchy}. Unfortunately, such Siamese and triplet strategies block the synergy coming from multitask learning, since they can only model one similarity at a time.
Instead, we take inspiration from Movshovitz-Attias~\etal~\cite{movshovitz2017no} to learn our embedding, who introduced proxies as an alternative to a triplet strategy for instance retrieval. We propose a cooperative proxy-based loss, where instance, attribute and category entities have their own proxy derivations, corresponding to their respective level of semantic representation in the embedding.

Our main contribution is a \textit{cooperative embedding} where instance, attribute and category simultaneously cooperate to build a single image representation of multiple similarities.
First, we introduce an algebraic structure to build the cooperative embeddings.
Second, we derive a cooperative proxy-based loss to learn image similarities from multiple entities simultaneously.
Third, we show the benefits of the cooperative embeddings on multiple image retrieval tasks, as well as its ability to discover style evolution between- and within-categories by navigating through instances and attribute values.
Experiments are held on datasets of \textit{car models}~\cite{7299023} and extended \textit{fashion products}~\cite{liuLQWTcvpr16DeepFashion}.

\section{Related Work}
Over the years, the theory of categorization~\cite{murphy2004big} has stimulated the computer vision community to go beyond category naming.
Prototypes~\cite{rosch1978principles} were discovered to learn a subset of representations for object recognition~\cite{quattoni2008transfer}, or a mapping to the entry-level of categorization~\cite{ordonez2015predicting}.
Exemplar images were leveraged to learn associations among images for object recognition~\cite{frome2007image,malisiewicz2008recognition}.
Attributes described categories for zero-shot recognition based on expert knowledge~\cite{farhadi2009describing, lampert2014attribute, finegrained}, or non-semantic features~\cite{sharmanska2012augmented,yu2013designing}.
Instance representations were augmented with categories~\cite{gordoa2012leveraging} or attributes~\cite{douze2011combining,tao2015attributes,liuLQWTcvpr16DeepFashion} for image retrieval.
We follow this tradition and study the relations among instances, attributes and categories as the key ingredients of image retrieval.

Several datasets have been released to provide multiple types of object labels~\cite{finegrained,krause20133d,lampert2014attribute,huang2015cross,7299023,songCVPR16,liuLQWTcvpr16DeepFashion,han2017automatic,Ak_2018_CVPR,mcauley2015image,han2017learning,6248071,huang2015cross,hadi2015buy}. Nevertheless, to address the multi-similarity nature of images some issues remain to be tackled.
Some datasets are no longer publicly available~\cite{huang2015cross,6248071,hadi2015buy}; or contain noisy labels~\cite{liuLQWTcvpr16DeepFashion}; or miss either the instance notion~\cite{lampert2014attribute,Ak_2018_CVPR,finegrained}, the category notion~\cite{finegrained}, or the explicit attribute notions~\cite{krause20133d,songCVPR16,han2017automatic}; or have other orthogonal purposes such as cross-domain image retrieval for fashion products~\cite{huang2015cross,liuLQWTcvpr16DeepFashion,6248071,hadi2015buy}, or fashion compatibility~\cite{mcauley2015image,han2017learning}.
Rather than collecting a new dataset, we re-annotated the existing In-Shop Clothes dataset~\cite{liuLQWTcvpr16DeepFashion}. The new labels will be released to foster research on multiple image similarities. We then rely on the newly re-annotated dataset, as well as the CompCars dataset~\cite{7299023}, for our experiments.

When multiple notions of similarity are involved, specific approaches are desired to relate them. Multitask learning~\cite{caruana1997multitask} is a simple and attractive concept. Given a neural network, it creates one head for every type of labels while relying on a shared representation. Although this setting still remains the norm nowadays~\cite{chen2015deep,huang2015cross,han2017automatic}, it lacks an explicit metric to measure image similarity. A recent success for building an image similarity metric comes from the triplet loss~\cite{weinberger2009distance,schroff2015facenet}. Triplets of images are compared through relative measurements. Nonetheless, several challenges emerge when training a triplet network, especially in the context of learning multiple image similarities.

To overcome the single-notion measure of the triplet loss, model architectures have to be adapted to handle the multi-similarity nature of images.
Liu~\etal~\cite{liuLQWTcvpr16DeepFashion} extended a multitask convolutional network (ConvNet) with a triplet loss to build a notion of instances in the embedding space.
Zhao~\etal~\cite{zhao2017memory} introduced a memory module and a multi-stage training for classifying and measuring similarities among categories and attributes. 
Ak~\etal~\cite{Ak_2018_CVPR} learned individual localization-aware heads for every attribute and category.
Alongside these efforts, other works tried to simplify these complex model architectures.
Veit~\etal~\cite{veit17conditionalsimilarity} introduced subspace embeddings to encode distinct attribute notions.
Zhao~\etal~\cite{zhao2018modulation} designed a modulation module for learning embeddings of binary attribute values.
Yet, these architectures don't cope with instance, attribute and category notions together.
We develop a cooperative structure to learn them together.

\begin{figure*}
\centering
\begin{subfigure}{.33\textwidth}
  \centering
  \caption{$\mathcal{L}_{Ins}$ in superspace}
  \includegraphics[width=0.80\linewidth]{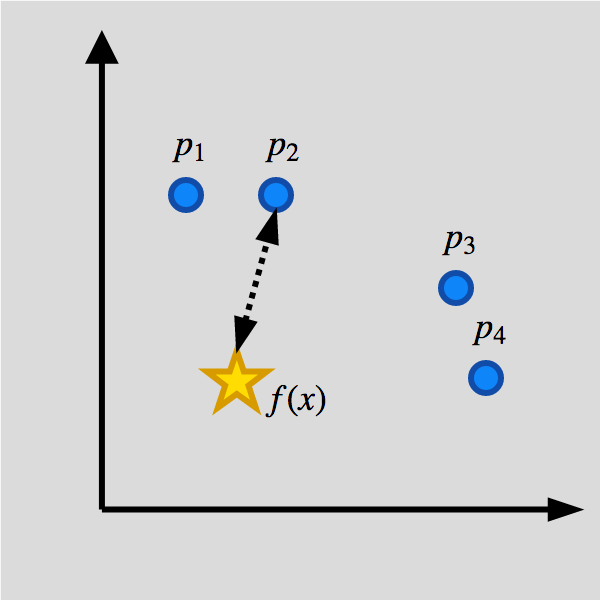}
  \label{fig:sfig1}
\end{subfigure}\hfill
\begin{subfigure}{.33\textwidth}
  \centering
    \caption{$\mathcal{L}_{Attr_1},\mathcal{L}_{Attr_2}$ in subspace 1, 2}
  \includegraphics[width=0.80\linewidth]{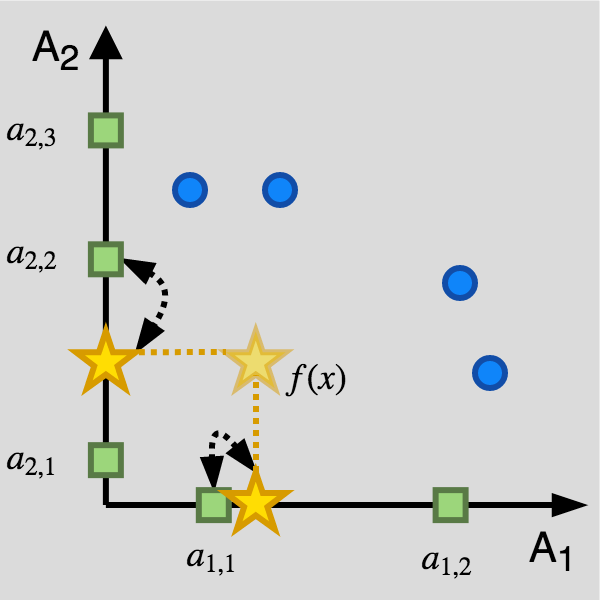}
  \label{fig:sfig2}
\end{subfigure}\hfill
\begin{subfigure}{.33\textwidth}
  \centering
    \caption{$\mathcal{L}_{Cat}$ in superspace}
  \includegraphics[width=0.80\linewidth]{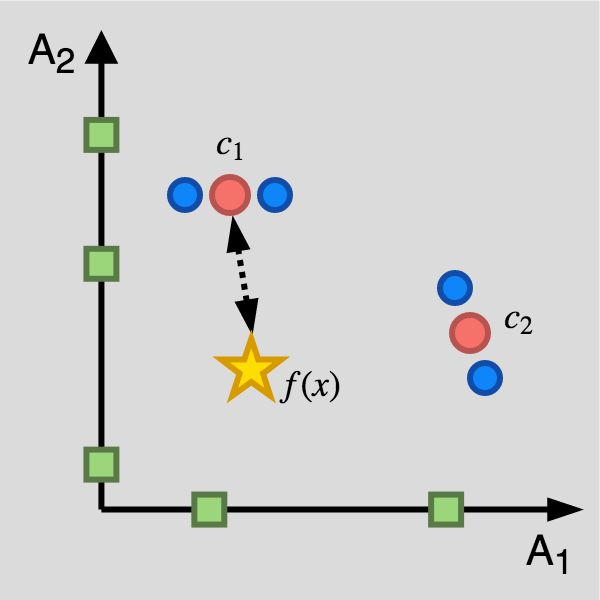}
  \label{fig:sfig3}
\end{subfigure}
\begin{subfigure}{.8\textwidth}
  \centering
  \vspace{10px}
  \includegraphics[width=\linewidth]{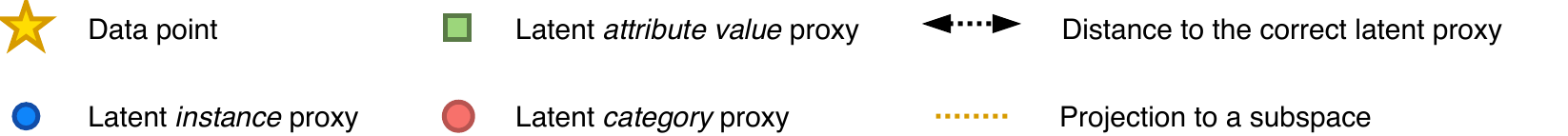}
  \vspace{-10px}
  \label{fig:sfig4}
\end{subfigure}
\caption{\textbf{2D toy example of the cooperative proxies for multiple similarities.}
(a) First, we learn an embedding space that revolves around instances.
$\mathcal{L}_{Ins}$ encourages images in the superspace to be close to their correct latent \textit{instance} proxy (Eq.~\ref{eq:pi}).
(b) Secondly, we enforce dimensions of the embedding space to encode separate attribute representations. Here, the axes encode attribute $A_1$ and $A_2$.
$\mathcal{L}_{Attr_k}$ encourages images to be close to their correct latent \textit{attribute value} proxy in their respective subspace (Eq.~\ref{eq:pv}).
(c) Third, we derive a meta-representation from the instances, which corresponds to a category-related notion.
$\mathcal{L}_{Cat}$ encourages images in the superspace to be close to their correct latent \textit{category} proxy (Eq.~\ref{eq:pc}).
}
\label{fig:emb}
\end{figure*}

To circumvent the complex training scheme of the triplet loss~\cite{schroff2015facenet,songCVPR16}, the idea of learning an embedding with prototypes has emerged.
Mensink~\etal.~\cite{mensink2013distance} introduced the nearest class mean classifier for few-shot learning. A linear projection mapped the features and pre-computed category centers to a joint embedding  using a probabilistic model.
Wen~\etal~\cite{wen2016discriminative} proposed to update the centers along with the network parameters for face recognition. To prevent their training from collapsing, they adjoined a cross-entropy loss.
Center representations can also be relaxed by learning an approximate representation, referred as a proxy.
Movshovitz-Attias~\etal~\cite{movshovitz2017no} introduced proxies to the neighbourhood component analysis~\cite{goldberger2005neighbourhood} for instance retrieval.
Still, center- or proxy-based losses are also restricted to the learning of one single similarity at a time. 
To remedy that shortcoming, we incorporate the proxy-based loss into the cooperative embeddings to cope with the different levels of the composition and the multiple notions of similarity.


\section{Model}

\subsection{Cooperative embeddings} \label{ssec:comp}
We represent visual similarities in a Euclidean embedding space $U \in \mathbb{R}^N$. A set of Euclidean subspaces $\{A_1, \ldots, A_K\}$ composes the embedding space $U$. Every subspace $A_k \in \mathbb{R}^n$ in the set refers to an attribute $k$, where $n=N/K$ and $k=1,\ldots,K$ attributes. We get an ordered sequence of attribute subspaces $<A_1, \ldots, A_K>$, which concatenates the subspaces in a specific order to form the superspace $U$. Hence, $U$ is the direct sum operation $\oplus$ of its subspaces $U = A_1 \oplus \ldots \oplus A_K$.

This algebraic structure of the embedding space gives rise to the following roles. First, instances lie in the superspace. They are composed of the direct sum of all attribute subspaces. Second, attributes are axis-aligned and correspond to distinct subspaces. Specific dimensions in the superspace encode different attribute notions. Third, categories also lie in the superspace. Their representation is the average of all the instance representations they comprise.

The embedding function $f(\cdot)$ learns to map an image $x$ to the embedding space $U$.
$f(\cdot)$ is typically a ConvNet for feature extraction followed by a linear projection to the embedding space. We rely on a proxy-based formulation to learn every similarity notion in the superspace and all subspaces. Latent proxies are initialized randomly and are simply considered as model parameters. During training, gradients are backpropagated to update both model parameters and all latent proxies simultaneously.

\subsection{Cooperative proxy-based loss} \label{ssec:loss}

\noindent\textbf{Instance loss.}
Every instance is represented by a latent instance proxy $p_i \in U$ with $i=1, \dots, I$ instances (Fig.~\ref{fig:sfig1}). In other words, we have a distributional model of meaning where instances are represented by a continuous vector representation.
We minimize the Euclidean distance between $f(x)$ and $p_i$ through a probabilistic model. The loss function $\mathcal{L}_{Ins}$ is then a softmax embedding loss for instance learning in the superspace:
\begin{align}
\label{eq:pi}
  \mathcal{L}_{Ins} = -\log\frac{\exp(-D(f(x), p_{i}))}{\sum_{z \in \mathbb{Z}_I}\exp(- D(f(x), p_{z})) },
\end{align}
where $D(f(x), p_{i}) = \|f(x) - p_{i}\|^2_2$ is the Euclidean distance between the projection of an image $f(x)$ and the latent instance proxy $p_{i}$ of the instance $i$ in the embedding space $U$. $\mathbb{Z}_I$ denotes the set of all the latent instance proxies. The softmax function is an essential component to learn the embedding space. Without a normalization scheme the model will learn a trivial solution by mapping both latent proxies and image projections to zero. 

\vspace{5pt}\noindent\textbf{Attribute loss.}
An attribute value $v$ of attribute $k$ is represented by a latent attribute value proxy $a_{k, v}$ in the subspace $A_{k}$ (Fig.~\ref{fig:sfig2}).
The projection $f(\cdot)$ of an image $x$ to the subspace $A_{k}$ is obtained by masking the projection $f(x)$ with a binary mask $M_k$ corresponding to the attribute $k$. This masking operation corresponds to $f_k(x) = M_{k} \odot f(x)$, where $\odot$ is the element-wise multiplication. The masks serve as an element-wise gating function to force the network to learn attribute-related notions in specific and separate dimensions of the embedding space $U$. Moreover, the masks are non-overlapping and different from each other. Note that if the dimension of every subspace is equal to one, \ie $n=1$, the concatenation of the ordered sequence $<M_1, \ldots, M_K>$ forms the standard basis of $U$.

The learning process relies on a probabilistic model to assign an image $x$ to its value $v$ of attribute $k$ using the softmax function over Euclidean distances in the subspace $A_k$. The loss function $\mathcal{L}_{Attr_k}$ is a softmax embedding loss for attribute $k$ learning in its respective subspace:
\begin{align}
\label{eq:pv}
  \mathcal{L}_{Attr_k} = -\log \frac{\exp(- D_{M_{k}}(f(x), a_{k, v}))}{\sum_{z \in \mathbb{Z}_{A_k}}\exp(-D_{M_{k}}( f(x), a_{k, z}))},
\end{align}
where $D_{M_{k}}(f(x), a_{k, v}) = \| M_{k} \odot f(x) - M_{k} \odot a_{k, v} \|^2_2$ is the Euclidean distance between the projection of an image $f(x)$ and the latent attribute value proxy $a_{k, v}$ of the attribute value $v$ in the subspace $A_{k}$. $\mathbb{Z}_{A_k}$ denotes the set of all the latent attribute value proxies of attribute $k$. Every  subspace then learns a separate representation.

The above formulation considers attribute values to be categorical. Nevertheless, some attribute values can be ordered by count or by a real value. This can be incorporated in our formulation by defining an \textit{a priori} proximity matrix $\mathbf{P}_k$ to order all latent proxies $a_{k,v}$ within the subspace of attribute $k$. For convenience, we drop the subscript $k$.
We build $\mathbf{P}$ with a Gaussian similarity kernel:
\begin{align}
\label{eq:p}
  P_{vu} = \exp \bigg( -\frac{\lVert r_{a_{v}} - r_{a_{u}} \rVert^2_2}{2\sigma^2} \bigg),
\end{align}
\noindent where $r_{a_{v}}$ is the rank of the value $v$ of an attribute $k$. A regularizer $R_{Order}$ is then added to the loss during training to encourage an ordering of the latent proxies per subspace:
\begin{align}
\label{eq:r}
  R_{Order} = \lVert \mathbf{S} - \mathbf{P} \rVert_F,
\end{align}
\noindent where $\mathbf{S}$ is the cosine similarity matrix of all latent proxies $a_{v}$ of an attribute $k$. Note that this regularizer considers latent proxies rather than all image embeddings.

\vspace{5pt}\noindent\textbf{Category loss.}
A latent category proxy $c_y$ is defined as the average of all latent instance proxies of category $y$ (Fig.~\ref{fig:sfig3}). The intuition is that categories emerge when grouping similar instances. Consider the set $\mathbb{Y}$ of all instances of the category $y$, then $c_y = \frac{1}{|\mathbb{Y}|}\sum_{i \in \mathbb{Y}}p_i$.

Similarly, the learning process relies on a probabilistic model to assign an image $x$ to its category $y$ by applying a softmax function over Euclidean distances in the superspace. The loss function for category learning $\mathcal{L}_{Cat}$ is the softmax embedding loss for categories in the superspace:
\begin{align}
\label{eq:pc}
  \mathcal{L}_{Cat} = -\log \frac{\exp(-D(f(x), c_{y}))}{\sum_{z \in \mathbb{Z}_C}\exp(- D(f(x), c_{z})) },
\end{align}
where $D(f(x), c_{y}) = \|f(x) - c_{y}\|^2_2$ is the Euclidean distance between the projection of an image $f(x)$ and the latent category proxy $c_{y}$ of the ground-truth category label $y$ in the embedding space $U$. $\mathbb{Z}_C$ denotes the set of all latent category proxies. Minimizing Eq.~\ref{eq:pc} then influences the learning of the latent instance proxies.

\vspace{5pt}\noindent\textbf{Cooperative loss.}
The final loss spans all levels of the cooperative embeddings. It minimizes a weighted sum of the instance and category losses in the superspace and the attribute losses in all the distinct subspaces:
\begin{align}
\label{eq:ica}
  \mathcal{L}_{CE} = \lambda_{Ins} \mathcal{L}_{Ins} + \frac{\lambda_{Attr} }{K}\sum_k\mathcal{L}_{Attr_k} + \lambda_{Cat} \mathcal{L}_{Cat},
\end{align}
where $\lambda_{Ins}$, $\lambda_{Attr}$, and $\lambda_{Cat}$ denote trade-off hyperparameters to control the contribution of $\mathcal{L}_{Ins}$, $\mathcal{L}_{Attr_k}$ and $\mathcal{L}_{Cat}$, respectively. At every iteration, the supervision of the model takes as input the instance $i$, the set of exhibited attributes and their values, as well as the category $y$.
Contrary to the single similarity learning of triplet strategies, the proposed loss function handles all labels at the same time to learn multiple similarity measures simultaneously during training.
Note that some images might not express all attributes $K$ defined in the dataset, \eg a \textit{skirt} doesn't have a \textit{sleeves length} attribute. In this case, the contribution of the missing attribute in Eq.~\ref{eq:ica} is ignored.

\vspace{5pt}\noindent\textbf{Embedding regularization.} We prevent the embedding representations from exploding via an L2 regularization term on the embedding space:
\begin{align}
\label{eq:final}
  \mathcal{L}=\mathcal{L}_{CE} + \lambda_{Reg} \|f(x)\|^2_2,
\end{align}
where $\lambda_{Reg}$ is a hyperparameter to control the amount of L2 regularization. Such regularization promotes a better generalization performance~\cite{hariharan2017lowshot}. It prevents the representations $f(\cdot)$ to drift away from latent proxies, which forces the model to encode more useful semantic information~\cite{veit17conditionalsimilarity}.

\begin{figure}[t]
\centering
\includegraphics[width=0.9\linewidth]{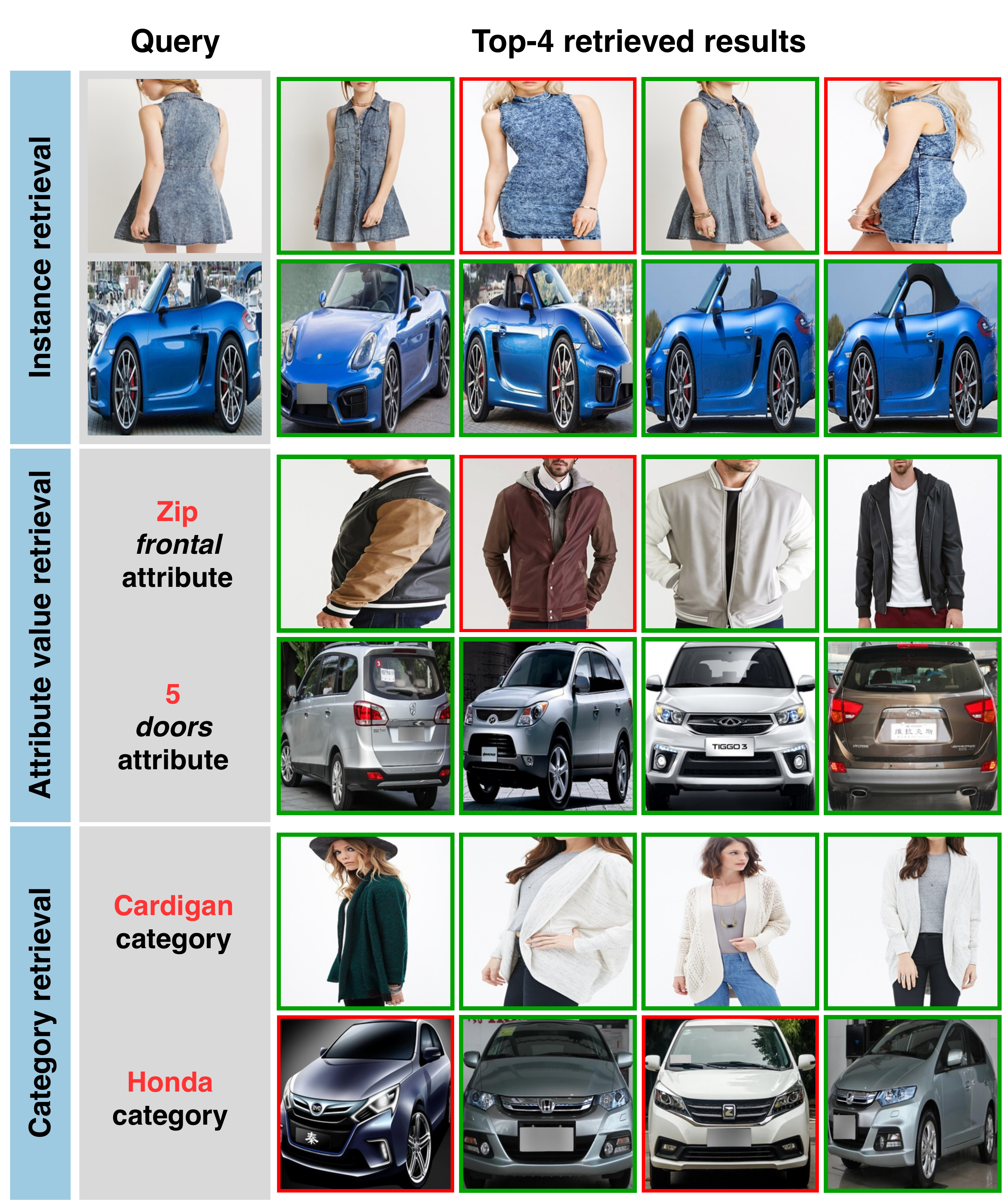}
\caption{\textbf{Example queries and their top-4 retrieved images} on the In-Shop Clothes-8 (odd rows) and CompCars (even rows) datasets. Correct matches are marked in green and incorrect matches are in red.
}
\label{fig:examples}\vspace{-10px}
\end{figure}

\section{Experimental Setup}

\subsection{Datasets} \label{ssec:dataset}
\noindent\textbf{In-Shop Clothes-8} is a fashion dataset based on the original In-Shop Clothes dataset~\cite{liuLQWTcvpr16DeepFashion}. We manually re-annotated the category and attribute labels for every instance of the dataset\footnote{The new labels will be released publicly.}. The whole protocol is described in the supplementary materials.
The original In-Shop Clothes dataset provides a large number of clothing products along with multiple views and a rich description of several sentences. However, the labeling was done in a weakly-supervised manner. This creates scarce attribute values, duplicates and incoherencies~\cite{2018arXiv180711674Z}.
The In-Shop Clothes-8 dataset defines the instances as the \textit{product id} and categories as the \textit{clothes category} (12 different). We specify 8 new attributes (for a total of 59 attribute values): 6 \textit{fabric}, 7 \textit{frontal features}, 6 \textit{hemline}, 13 \textit{neckline}, 15 \textit{prints}, 4 \textit{shoulder line}, 6 \textit{sleeves length} and 2 \textit{silhouette} values.
We keep the same splits as in~\cite{liuLQWTcvpr16DeepFashion}. After cleaning and relabeling, the train split contains 25,862 images from 3996 unique fashion products and the test split 26,797 images from 3982 separate and unique fashion products. The test split is further partitioned in a query and a gallery splits of 14,201 and 12,596 images, respectively.

\noindent\textbf{CompCars} is a car dataset of 78,126 images~\cite{7299023} with diverse backgrounds and multiple viewpoints. The dataset provides annotations about the model, the make and specific characteristics of the cars. We define the instances as the \textit{car model} and the categories as the \textit{car make} (44 shared between training and testing sets). Additionally, we are interested in the 3 provided attributes (for a total of 20 attribute values): 4 \textit{number of doors}, 4 \textit{number of seats} and 12 \textit{car type} values.
Following~\cite{7299023}, we train on the first split of 30,955 images from 431 unique car models and test on the second split of 4,454 images from 111 separate and unique car models. 8 car models were removed from the training set because no labels were available.

\subsection{Evaluation metrics}
We evaluate the generalization capability of our model to retrieve images of unseen instances in a test gallery.
Three retrieval tasks are performed (Figure~\ref{fig:examples}):

\noindent\textbf{Instance retrieval.} Given a query image, this query by example experiment finds the nearest neighbor images in the test gallery of unseen instances.
We report the recall at $K$ (R@$K$)~\cite{Manning2008,jegou2011product}, which evaluates the recall of the $K$ retrieved images. If at least one image of the same instance appears in the $K$ nearest neighbours, it is considered as a hit. We use $K=1$ in all experiments.

\noindent\textbf{Attribute value retrieval} and \textbf{Category retrieval.} Given a query term, these query retrieval experiments find the nearest neighbor images in the test gallery of unseen instances. The query term is built by averaging all the training images that exhibit the intended term,~\eg the query for \textit{zip frontal} attribute corresponds to the averaged features of all training images of fashion clothes with a \textit{zip frontal} attribute. Note that an attribute value query spans only its corresponding attribute subspace while a category query spans the superspace. We report the mean average precision (mAP)~\cite{Manning2008}.

\subsection{Implementation details}  \label{ssec:details}
The backbone network rests on ResNet50~\cite{he2016deep}, pre-trained on ImageNet~\cite{ILSVRC15}. To produce the embedding space, the classification layer is removed and replaced by a linear projection initialized with random weights. Latent proxies are also initialized with random weights. The model minimizes the loss function described in Eq.~\ref{eq:final} using the Adam stochastic optimizer algorithm~\cite{kingma2014adam} (minibatch size of 128, learning rate of $1\times10^{-4}$, $\beta_1=0.9$, $\beta_2=0.999$, weight decay of $5\times10^{-5}$). We set $\lambda_{Ins}=\lambda_{Cat}=\lambda_{Attr}=1$. Fine-tuning of ResNet operates at a learning rate 10$\times$ smaller, while the updates of the latent proxies operate at a learning rate 10$\times$ bigger. Images are cropped given their bounding box labels and resized to $224\times224$. During training, horizontal flipping is applied. In the image retrieval tasks, L2 normalization is applied per subspace. Training and evaluation of baselines follow the same process.
\setlength\dashlinegap{5pt}
\setlength\dashlinedash{2pt}
\begin{table}
\begin{subtable}[t]{\linewidth}
\centering
\resizebox{0.85\columnwidth}{!}{
\begin{tabular}{@{}cccccc}
\specialrule{.1em}{.05em}{.05em}
$\lambda_{Ins}$ &  $\lambda_{Attr}$ & $\lambda_{Cat}$ & Instance & Attribute & Category \\
                &           &        & (R@1)    & (mAP)    & (mAP) \\
\cmidrule(lr){1-3} \cmidrule(lr){4-6} 
\ding{51} &  &  & 81.01 & 35.53 & 55.93 \\
 & \ding{51} &  & 14.28 & 38.00 & 36.20 \\
 &  & \ding{51} & 2.29 & 16.37 & 83.93 \\
\cdashline{1-6}\noalign{\vskip 0.5ex}
\ding{51} &  & \ding{51} & 79.25 & 33.20 & 81.35 \\
  & \ding{51} & \ding{51} & 22.87 & 39.07 & \textbf{84.44} \\
 \ding{51} & \ding{51} &  & \textbf{81.99} & 42.52 & 63.98 \\
 \ding{51} & \ding{51} & \ding{51} & 78.12 & \textbf{43.18} & 81.70 \\
\specialrule{.1em}{.05em}{.05em}
\end{tabular}
}
\caption{\textbf{Composing the embedding} (In-Shop Clothes-8 dataset).
Instance, attribute subspaces, and category settings in isolation (rows 1-3). Additional settings (rows 4-7) enabled by our cooperative embedding showing how entities complement each other. The final structure yields the best overall performance.}
\label{tab:tradeoff}
\end{subtable}

\medskip

\begin{subtable}[t]{\linewidth}
\centering
\resizebox{0.85\columnwidth}{!}{
\begin{tabular*}{\textwidth}{@{\extracolsep{\fill}}cccc}
\specialrule{.1em}{.05em}{.05em}
\mc{Subspace} & Instance & Attribute & Category \\
 \mc{width $n$} & (R@1)    & (mAP)    & (mAP) \\
\cmidrule(lr){1-1} \cmidrule(lr){2-4}
\mc{1}   & 3.95  & 8.23 & 53.77 \\
\mc{5}   & 53.20 & 24.54 & 77.51 \\
\mc{25}  & 76.35 & 40.97 & 81.02 \\
\mc{50}  & \textbf{78.12} & 43.18 & 81.70 \\
\mc{100} & 78.05  & \textbf{44.05} & \textbf{81.71} \\
\specialrule{.1em}{.05em}{.05em}
\end{tabular*}
}
\caption{\textbf{Wide \vs Narrow subspaces} (In-Shop Clothes-8 dataset). Performance only slightly improves after $n=50$.}
\label{tab:dim}
\end{subtable}

\medskip

\begin{subtable}[t]{\linewidth}
\centering
\resizebox{0.85\columnwidth}{!}{
\begin{tabular*}{\textwidth}{@{\extracolsep{\fill}}ccccc}
\specialrule{.1em}{.05em}{.05em}
\mc{Ordering} & \multicolumn{2}{c}{\# Doors} & \multicolumn{2}{c}{\# Seats} \\
 & MAE $\downarrow$ & MRR $\uparrow$ & MAE $\downarrow$ & MRR $\uparrow$ \\
\cmidrule(lr){1-1} \cmidrule(lr){2-5}
 & 0.372 & 0.717 & 0.356 & 0.528\\
\mc{\ding{51}} & \textbf{0.365} & \textbf{0.723} & \textbf{0.341} & \textbf{0.561} \\
\specialrule{.1em}{.05em}{.05em}
\end{tabular*}
}
\caption{\textbf{Non-ordered \vs ordered latent proxies} (CompCars dataset). Ordered attributes can be modeled with an \textit{a priori} regularizer.}
\label{tab:order}
\end{subtable}
\caption{\textbf{Embedding ablation.}}
\label{tab:table1}\vspace{-10px}
\end{table}

\section{Results}

\subsection{Embedding ablation}
\noindent\textbf{Composing the embedding.} 
We investigate the influence of the cooperative structure on the In-Shop Clothes-8 dataset. We vary the trade-off hyperparameters of the loss (Eq.~\ref{eq:ica}) to control the arrangement of the embedding in Table~\ref{tab:tradeoff}. 
Combinations produce higher scores than isolated embeddings (rows 1-3), which shows the benefits coming from integrating multiple entities (rows 4-7). Nevertheless, integrating three entities remains challenging as they belong to different levels of semantic representations. A complementary duality exists between \textit{attributes} and the other two entities: \textit{attributes} are a mid-level representation but always help improve the scores of \textit{instance} and \textit{category} retrieval. A competing duality appears between \textit{instances} and \textit{categories}: focusing on \textit{categories} pushes the embedding to be agnostic to \textit{instances} differences. Adding the category entity slightly dampens instance retrieval, and vice versa. This may also originate from atypical instances at the edge of category boundaries we discover in Sec.~\ref{sec:style}.
Integrating instance, attribute and category yields the best overall scores.

\noindent\textbf{Wide \vs Narrow subspaces.} We study the influence of the number of dimensions $n$ per subspace  on the In-Shop Clothes-8 dataset in Table~\ref{tab:dim}. The more dimensions per subspace, the better the performance.
We choose $n=50$ as a good compromise between performance and compactness. For the In-Shop Clothes-8 dataset, this then results in a superspace of size $N=400$, with $K=8$ attribute subspaces of size $n=50$ each.

\noindent\textbf{Non-ordered vs Ordered latent proxies.}
We evaluate the regularization term in Eq.~\ref{eq:r} on two ordered and countable attributes on the CompCars dataset in Table~\ref{tab:order}.
The \textit{number of doors} and the \textit{number of seats} differ from categorical attributes because their values can be ordered by count.
We report the mean absolute error (MAE) and the mean reciprocal rank (MRR) between the predicted attribute value and the true one.  
We observe a noteworthy MRR improvement for the \textit{number of seats} while others show slightly improved scores.
We only notice a negligible drop in the other retrieval metrics (\ie R@1 and mAP scores).
If needed, an ordering can be induced in our formulation.

\begin{table}
\begin{subtable}[t]{\linewidth}
\centering
\resizebox{0.85\columnwidth}{!}{
\begin{tabular}{ccccc}
\specialrule{.1em}{.05em}{.05em}
Instance & Attribute & Instance  & Attribute & Category \\
proxies & proxies & (R@1)    & (mAP)    & (mAP) \\
\cmidrule(lr){1-2} \cmidrule(lr){3-5}
Fixed & Fixed & 68.07  & 33.59 & 64.56 \\
Learned & Fixed & 77.46 & 36.99 & 81.26 \\
Fixed & Learned & 69.13 & 40.96 & 69.60  \\
\cdashline{1-5}\noalign{\vskip 0.5ex}
Learned & Learned & \textbf{78.12} & \textbf{43.18} & \textbf{81.70} \\
\specialrule{.1em}{.05em}{.05em}
\end{tabular}
}
\caption{\textbf{Learned \vs Fixed latent proxies} (In-Shop Clothes-8 dataset). Fixing the latent proxies dampens the performance by restricting the model to shape both the superspace and the subspaces.
}
\label{tab:fixed}
\end{subtable}

\medskip

\begin{subtable}[t]{\linewidth}
\centering
\resizebox{0.85\columnwidth}{!}{
\begin{tabular}{ccccc}
\specialrule{.1em}{.05em}{.05em}
Corruption & Mode & Instance  & Attribute & Category \\
level & & (R@1)    & (mAP)    & (mAP) \\
\cmidrule(lr){1-1} \cmidrule(lr){2-2} \cmidrule(lr){3-5}
40\% & Absence & 77.66 & 42.50 & 81.50 \\
& Swap   & 76.95  & 36.99 & 80.87 \\
& Both  & 77.21 & 37.57 & 80.99  \\
\cdashline{1-5}\noalign{\vskip 0.5ex}
20\% & Absence & 77.40 & 42.52 & 81.16  \\
 & Swap & 77.59 & 39.24 & 81.15  \\
 & Both & 77.77 & 39.28 & 81.65  \\
 \cdashline{1-5}\noalign{\vskip 0.5ex}
 0\% & n/a & \textbf{78.12} & \textbf{43.18} & \textbf{81.70} \\
\specialrule{.1em}{.05em}{.05em}
\end{tabular}
}
\caption{\textbf{Attribute labels corruption} (In-Shop Clothes-8 dataset). Swapping the attribute values hurts the performance more than having an absence of attribute values.}
\label{tab:noise}
\end{subtable}

\medskip
\begin{subtable}[t]{\linewidth}
\centering
\resizebox{0.86\columnwidth}{!}{
\begin{tabularx}{\linewidth}{@{\extracolsep{\fill}}cccc}
\specialrule{.1em}{.05em}{.05em}
\mc{$\lambda_{Reg}$} & Instance & Attribute & Category \\
  & (R@1)    & (mAP)    & (mAP) \\
\cmidrule(lr){1-1} \cmidrule(lr){2-4}
\mc{0}                   & 76.82 & 41.61 & 80.97 \\
\mc{$5 \times 10^{-3}$}  & 76.93 & 41.58 & 81.05 \\
\mc{$5 \times 10^{-2}$}  & 76.91 & 41.67 & 81.12 \\
\mc{$5 \times 10^{-1}$}  & \textbf{78.12} & \textbf{43.18} & \textbf{81.70} \\
\specialrule{.1em}{.05em}{.05em}
\end{tabularx}
}
\caption{\textbf{Small \vs Strong feature regularization} (In-Shop Clothes-8 dataset). Strong feature regularization improves the performance.}
\label{tab:ereg}
\end{subtable}

\caption{\textbf{Loss function ablation.}}
\label{tab:table2}\vspace{-7px}
\end{table}

\begin{figure*}
\centering
  \includegraphics[width=0.95\linewidth]{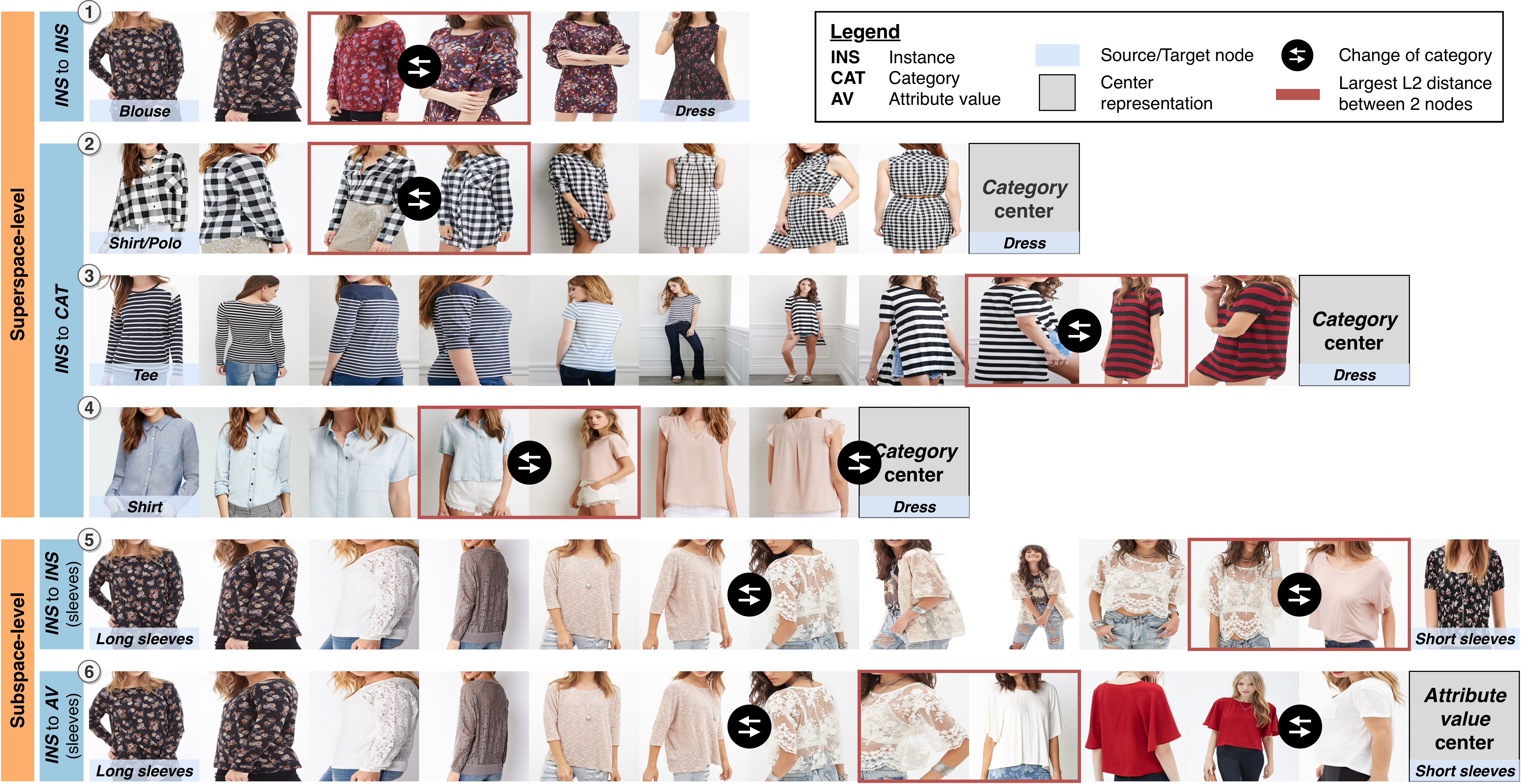}
  \caption{\textbf{Visual transitions between two categories} from a source image to a target image or a target representation. The red frame indicates the largest L2 distance between two images in the path, which usually coincides with a change of category.
}
  \label{fig:trans}\vspace{-5px}
\end{figure*}

\subsection{Loss function ablation}
\noindent\textbf{Learned \vs Fixed latent proxies.} We evaluate the influence of learning the latent proxies  on the In-Shop Clothes-8 dataset in Table~\ref{tab:fixed}.
Fixing the proxies means that we don't take the partial derivative of Eq.~\ref{eq:final} \wrt the latent proxies to update them during training.
Latent instance proxies $p_i$ and latent attribute value proxies $a_{k,v}$ can be \textit{learned} or \textit{fixed}.
Latent category proxies $c_y$ are derived from $p_i$, so both share the same learning state.
When both instance and attribute proxies are fixed, it yields the lowest scores. The model maps to a fixed embedding space without the possibility to shape it.
As soon as one is learned, an improvement is observed on all three metrics.
When both are learned, it provides the highest scores. The model can learn its weights and shape the embedding to be more semantically meaningful.
Thus, learning latent proxies improves generalization.

\noindent\textbf{Attribute labels corruption.} We study the effect of noisy attribute value labels  on the In-Shop Clothes-8 dataset in Table~\ref{tab:noise}. Most of the re-annotation effort focused on removing incoherencies in the attribute labels.
For example, there were items labeled as \textit{long sleeves} and \textit{sleeveless} at the same time. This makes attribute and category retrieval impossible with the original labels.
Instead, we simulate noisy labels by corrupting one attribute value per instance for 20\% and 40\% of the instances randomly. The corruption sets an \textit{absence} of a value, a \textit{swapping} to another value, or a combination of \textit{both}.
Only the attribute retrieval scores are affected since these values were corrupted.
When attribute values are absent, the model still performs surprisingly well. This comes from the proxy-based formulation of the loss function which excludes their contribution (see Eq.~\ref{eq:ica}).
With a triplet loss, an image with an absent value can be incorrectly sampled as a negative sample.
When attribute values are swapped, this reduces the performance by introducing label inconsistencies.
Hence, having correct attribute value labels matters more than an absence in our model.

\begin{figure}
\centering
  \includegraphics[width=0.8\linewidth]{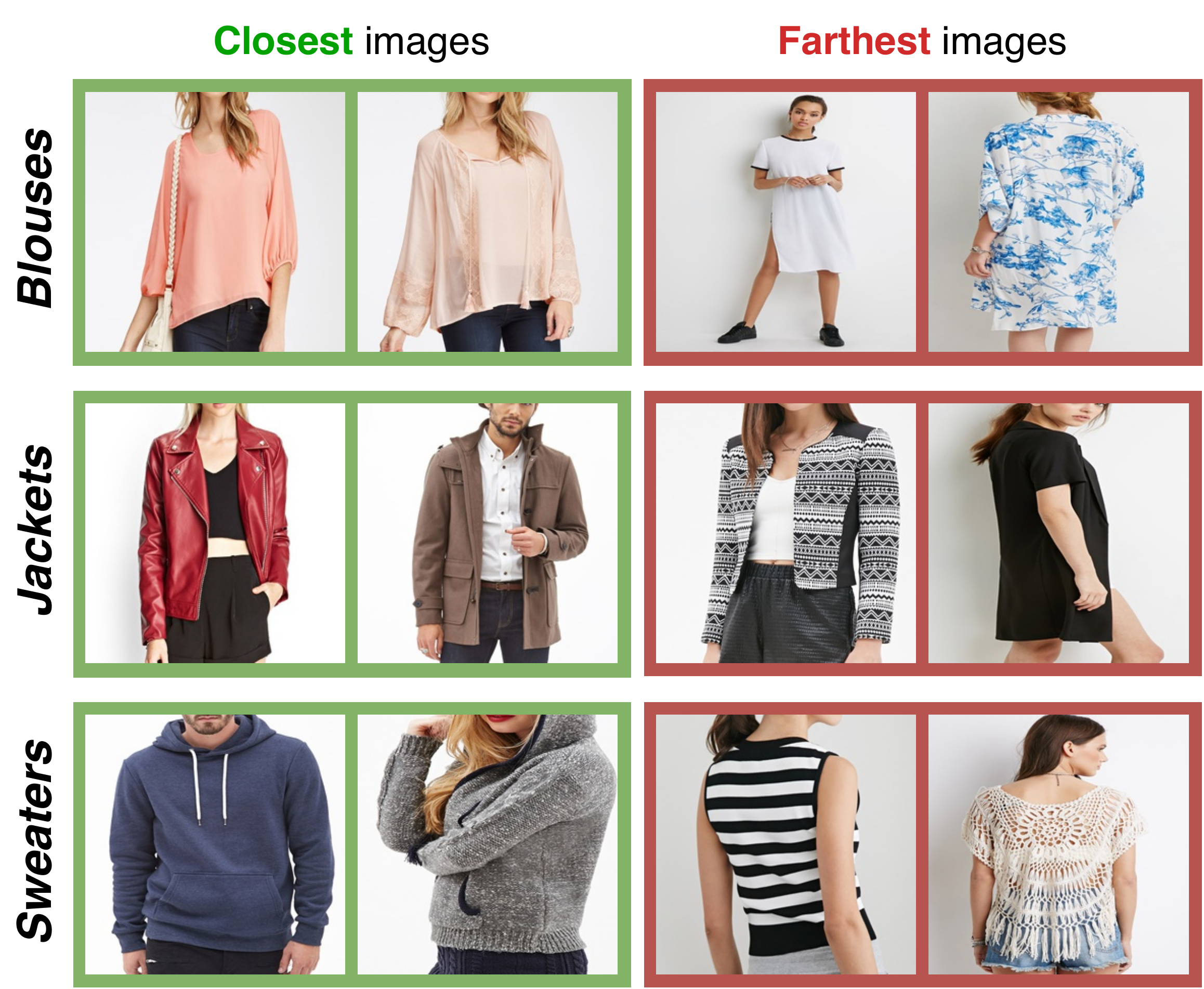}
  \caption{\textbf{Visual transitions within a category} in the In-Shop Clothes-8 dataset. Green boxes depict the closest images to the empirical category center (typical) while the red boxes the farthest images (atypical).}
  \label{fig:atyp}\vspace{-10px}
\end{figure} 

\begin{table*}
\centering
\resizebox{2\columnwidth}{!}{
\begin{tabular}{lcccccccccc}
\specialrule{.1em}{.05em}{.05em}
& \multicolumn{5}{c}{\textbf{CompCars}} &  \multicolumn{5}{c}{\textbf{In-Shop Clothes-8}} \\
\cmidrule(lr){2-6} \cmidrule(lr){7-11}
          & Output & Subspaces & Instance &  Attribute  & Category & Output & Subspaces  & Instance  & Attribute & Category \\
                & size &  & (R@1)    & (mAP)    & (mAP) & size &  & (R@1)    & (mAP)    & (mAP) \\
\cmidrule(lr){1-1} \cmidrule(lr){2-2} \cmidrule(lr){3-3} \cmidrule(lr){4-6} \cmidrule(lr){7-7} \cmidrule(lr){8-8} \cmidrule(lr){9-11}
a. Multitask (MTL)~\cite{caruana1997multitask}$^\star$ 
& 2048 & n/a       & 90.97 & 53.74 & 22.17
& 2048 & n/a       & 56.34 & 39.42 & 74.56 \\
b. MTL + triplet loss~\cite{liuLQWTcvpr16DeepFashion}$^\star$ 
& 2048 & n/a       & \textbf{93.22}  & 55.14 & 24.65
& 2048 & n/a       & 64.49 & 36.38 & 68.98 \\
c. Concatenation~\cite{veit17conditionalsimilarity}$^\ddagger$
& 250  & 5        & 75.71 & 53.92 & 28.86 
& 500  & 10         & 55.05 & 35.99 & 77.40 \\
d. Modulation~\cite{zhao2018modulation}$^\ddagger$
& 5$\times$150 & n/a       & 82.40 & 50.74 & 26.72
& 10$\times$400 & n/a       & 53.80 & 29.14 & 71.36 \\
\cmidrule(lr){1-1} \cmidrule(lr){2-6} \cmidrule(lr){7-11}
\textit{This paper}
& 150  & 3         & 88.19 & \textbf{56.45} & \textbf{31.94} 
& 400  & 8         & \textbf{78.12}  & \textbf{43.18} & \textbf{81.70} \\
\specialrule{.1em}{.05em}{.05em}
\end{tabular}
}
\caption{
\textbf{Cooperation \vs alternatives for multiple similarities.}
All methods rely on the same ResNet50 backbone, image pre-processing pipeline, and on either our implementation$^\star$ or an adapted open source implementation$^\ddagger$.
Note that baselines~\cite{caruana1997multitask,liuLQWTcvpr16DeepFashion,veit17conditionalsimilarity,zhao2018modulation} were repurposed to handle instance, attribute and category retrieval tasks.
}
\label{tab:sota}\vspace{-10px}
\end{table*}

\noindent\textbf{Small \vs Strong feature regularization.} We assess the influence of the L2 regularization on the feature space on the In-Shop Clothes-8 dataset in Table~\ref{tab:ereg}. A small regularization doesn't influence the generalization performance significantly. When $\lambda_{Reg}=0.5$, it yields an improvement in all metrics. However, when $\lambda_{Reg}=1$, the regularization hurts the training process and doesn't converge. From now on, we then set $\lambda_{Reg}=0.5$.

\subsection{Discovering style evolution} \label{sec:style}
\noindent\textbf{Visual transitions between two categories.}
We find atypical instances in image transitions from one category to another by constructing a $k$-nearest neighbours graph~\cite{kaneva2010infinite,StatesAndTransformations}, with $k=5$.
Nodes correspond to images in the test gallery and edges to the L2 distance between two images.
The {Dijk}-stra's shortest path computes visual transitions between a source image and a target image (or empirical center representation).
Figure~\ref{fig:trans} shows visual transitions from the In-Shop Clothes-8 dataset.
In the superspace, visual transitions preserve most of the visual attribute properties (rows 1-4).
Surprisingly, largest L2 distances coincide with crossovers of category boundaries, depicting atypical instances.
Row 4 shows a failure case with multiple crossovers and no \textit{dress} images.
In subspaces, visual transitions become agnostic to the other attributes and the category. This explains why the \textit{sleeves} transition exhibits \textit{print} or \textit{fabric} attribute changes along with multiple category changes (rows 5-6).

\noindent\textbf{Visual transitions within a category.} We rank images in the test gallery \wrt the distance to their empirical category center~\cite{deselaers2011visual}. Figure~\ref{fig:atyp} shows three categories from the In-Shop Clothes-8 dataset. The closest images illustrate the typical instances of the category while the farthest images depict the images at the category boundary. Interestingly, they show atypical instances of the category. For example, in the \textit{blouse} category, some instances look like a \textit{dress} or a \textit{t-shirt}. Moreover, they can highlight potential issues, such as confusing viewpoints or remaining labeling error.
\vspace{-2px}
\subsection{Cooperation \vs alternatives}
\vspace{-2px}
We compare our method with 4 main alternatives in Table~\ref{tab:sota}. We repurpose the following works to make them handle instance, attribute and category similarities:\\
{\bf a. Multitask learning}:
A neural network learns to classify $K$ attributes in $K$ heads and categories in \textit{one} head, for a total of $K+1$ heads, as suggested by Caruana~\cite{caruana1997multitask}.\\
{\bf b. Multitask learning with a triplet strategy}: An additional triplet loss in the embedding space models instance similarity, following Liu~\etal~\cite{liuLQWTcvpr16DeepFashion}.\\
{\bf c. Concatenation with a triplet strategy}:
A concatenation of instance, attribute and category similarity notions forms a total of $K+2$ subspaces, as an extension of the $K$ attribute subspaces of Veit~\etal~\cite{veit17conditionalsimilarity}.\\
{\bf d. Modulation with a triplet strategy}:
A module transforms the features based on the notion of interest, as suggested by Zhao~\etal~\cite{zhao2018modulation}. While being very lightweight, it requires to store $K+2$ embedding spaces at inference time.

\noindent\textbf{Results on In-Shop Clothes-8.} Our cooperative embedding outperforms the other baselines in all three metrics. Building a metric space to cope with multiple notions is difficult. When adding a triplet loss related to the instance notion, a drop occurs in the other notions (b.). When creating separate subspaces with a triplet loss (c.), the model yields an average overall performance because the loss cannot handle all entities simultaneously.
The modulation module produces lower results than a concatenation as it was originally designed for binary attributes (d.).

\noindent\textbf{Results on CompCars.} Our cooperative embedding outperforms the other baselines in the attribute metric and produces slightly lower scores in the other two metrics. Overparametrized embedding spaces perform surprisingly well (a. and b.). This comes probably from the low variability in the images in the test gallery (see row 2 of Fig.~\ref{fig:examples}). Learning a much smaller embedding space with separate subspaces dampens the instance retrieval performance but improves attribute and category retrieval scores.

\vspace{-2pt}
\section{Conclusion}
\vspace{-2pt}
Based on the criss-crossing roles of instances, attributes and categories, we propose a \textit{cooperative} embedding to integrate them according to their level of semantic representation. Furthermore, we derive a \textit{cooperative} proxy-based loss to learn these three similarity notions simultaneously.
Experiments on datasets from two different domains show a better ability of the model to perform image retrieval of multiple entities compared to other alternatives.
We also explore the embedding space to discover intriguing instances between categories.
Failure cases in visual transitions are interesting because they show gaps product designers can exploit.
In the current form, attribute subspaces describe physical properties of objects but could also capture aesthetics, or cultural differences.

\noindent\textbf{Acknowledgements}.
The authors thank Mert Kilickaya, Pascal Mettes, Zenglin Shi, and Hubert Banville for useful discussions and feedback.
William Thong is partially supported by an NSERC scholarship.

{\small
\bibliographystyle{ieee}
\bibliography{arxiv}
}

\end{document}